\documentclass[sigconf]{acmart}
\usepackage[utf8]{inputenc}
\usepackage[english]{babel}
\usepackage{graphicx}
\usepackage{amsmath}
\usepackage{amsthm}
\usepackage{algorithm}
\usepackage{algorithmic}
\newcommand{\indep}{\rotatebox[origin=c]{90}{$\models$}}
\usepackage{subcaption}
\usepackage{bm}
\usepackage{booktabs}
\usepackage{enumitem}
\usepackage{cleveref}
\usepackage{multirow}
\usepackage{amsfonts}
\usepackage{xcolor,colortbl}

\definecolor{Gray}{gray}{0.85}

\newcolumntype{a}{>{\columncolor{Gray}}c}
\newcolumntype{b}{>{\columncolor{white}}c}
\newtheorem{theorem*}{Theorem}
\newcommand{\mat}[1]{{\bf #1}}   % matrix: bold
\newtheorem{assumption}{Assumption}

\AtBeginDocument{%
  \providecommand\BibTeX{{%
    \normalfont B\kern-0.5em{\scshape i\kern-0.25em b}\kern-0.8em\TeX}}}

\setcopyright{acmcopyright}
\copyrightyear{2022}
\acmYear{2022}
\setcopyright{acmcopyright}\acmConference[WSDM '22]{Proceedings of the Fifteenth ACM International Conference on Web Search and Data Mining}{February 21--25, 2022}{Tempe, AZ, USA}
\acmBooktitle{Proceedings of the Fifteenth ACM International Conference on Web Search and Data Mining (WSDM '22), February 21--25, 2022, Tempe, AZ, USA}
\acmPrice{15.00}
\acmDOI{10.1145/3488560.3498407}
\acmISBN{978-1-4503-9132-0/22/02}
\begin{document}
\fancyhead{}
\title{Causal Mediation Analysis with Hidden Confounders}

\author{Lu Cheng\textsuperscript{\rm 1},  Ruocheng Guo\textsuperscript{\rm 2}, Huan Liu\textsuperscript{\rm 1}}
\affiliation{\textsuperscript{\rm 1} School of Computing and Augmented Intelligence, Arizona State University, USA\\
\textsuperscript{\rm 2} School of Data Science, City University of Hong Kong, China}
\email{{lcheng35, huanliu}@asu.edu, ruocheng.guo@cityu.edu.hk}

\renewcommand{\shortauthors}{Cheng, et al.}

\begin{abstract}
An important problem in causal inference is to break down the total effect of a treatment on an outcome into different causal pathways and to quantify the causal effect in each pathway. For instance, in causal fairness, the total effect of being a male employee (i.e., treatment) constitutes its direct effect on annual income (i.e., outcome) and the indirect effect via the employee's occupation (i.e., mediator). Causal mediation analysis (CMA) is a formal statistical framework commonly used to reveal such underlying causal mechanisms. One major challenge of CMA in observational studies is handling \textit{confounders}, variables that cause spurious causal relationships among treatment, mediator, and outcome. Conventional methods assume sequential ignorability that implies all confounders can be measured, which is often unverifiable in practice. This work aims to circumvent the stringent sequential ignorability assumptions and consider \textit{hidden confounders}. Drawing upon proxy strategies and recent advances in deep learning, we propose to simultaneously uncover the latent variables that characterize hidden confounders and estimate the causal effects. Empirical evaluations using both synthetic and semi-synthetic datasets validate the effectiveness of the proposed method. We further show the potentials of our approach for causal fairness analysis.
\end{abstract}


\maketitle

\section{Introduction}
Consider the following two real-world problems.

\noindent\textbf{Example 1.} An E-commerce platform (e.g., Amazon) wants to help sellers promote products by introducing a new feature in the recommendation module in addition to users' organic search. However, with the two modules together contributing to the conversion rate, the improved performance of one module may render the effects of optimizing another module insignificant and sometimes even negative \cite{yin2019identification}. For instance, the new recommendation feature successfully suggests products that fulfill users' needs meanwhile notably curtails the user engagement in organic search. 

\noindent\textbf{Example 2.} Researchers in fair machine learning want to examine if female employees are systematically discriminated in various companies. They collect data from US census that record whether a person earns more than \$50,000/year, her age, gender, occupation, and other demographic information. A causal fairness analysis is conducted to estimate the effect of gender on the income: a nonzero effect might indicate discrimination. They further find out that the discrimination consists of the direct discrimination against female and the indirect discrimination against female through occupation. 

\begin{figure}
    \centering
      \includegraphics[width=.45\linewidth]{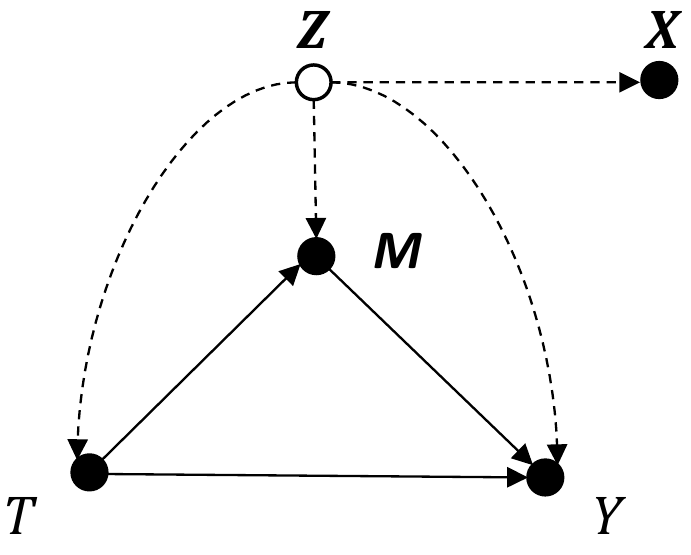}
    \caption{The causal diagram $G$ of CMA with proxy variables. $\mat{X}$ are the noisy proxies used to approximate HC ($\mat{Z}$).
    Solid dots and edges represent observed variables and causal relationships.}
\label{graphs}
\end{figure}

Underpinning the previous phenomena is a causal mediation analysis (CMA) where the total effect of a treatment (e.g., the new feature or gender) on the outcome (e.g., conversion rate or income) constitutes a \textit{direct causal effect}, e.g., gender $\rightarrow$ income, and an \textit{indirect causal effect} through the intermediate variable -- mediator, e.g., gender $\rightarrow$ occupation $\rightarrow$ income. CMA is a formal statistical framework aiming to quantify the direct effect and the indirect effect (causal mediation effect)\footnote{We use these two terms exchangeably in the rest of the paper.} of the treatment on the outcome. Despite its transparency and capability of fine-grained causal analysis, CMA confronts the conventional challenge when applied to observational studies: \textit{hidden confounders} (HC), a set of hidden variables $\mat{Z}$ that affects causal relationships among the treatment $T$, mediator, $\mat{M}$ and outcome $Y$ \cite{imai2010general}, as shown in Fig. \ref{graphs}.

Compared to standard causal inference, controlling for HC in CMA is even more challenging as confounders can lie on the causal pathways between any pair of variables among $T, \mat{M}, Y$. For example, in Fig. \ref{graphs}, $\mat{Z}$ can lie on the path $T\rightarrow \mat{M}$ or $\mat{M}\rightarrow Y$ or both. Therefore, research in CMA often has recourse to  the \textit{sequential ignorability assumption} \cite{imai2010general} which states that all confounders can be measured. This is clearly unverifiable in most observational studies, e.g., a user's preference in Example 1 or a person's sexual orientation in Example 2. Even gold-standard randomized experiments \cite{rubin2005causal} cannot provide us valid direct and indirect effect estimations in CMA \cite{keele2015causal} since we cannot randomize the mediator. 

In this paper, we seek to circumvent the sequential ignorability assumption by considering HC. Prior causal inference research \cite{griliches1986errors,pearl2012measurement} estimates total effect by imposing strong constraints on HC such as being categorical. Nevertheless, we cannot know the exact nature of HC (e.g., categorical or continuous), especially when HC can come from multiple sources induced by the mediator. To address this, we have recourse to the observed \textit{proxy variables} $\mat{X}$ \cite{angrist2008mostly} under the assumption that \textit{proxies and HC are inherently correlated}. For instance, we might approximate user preferences by measuring proxies such as her job type and zip code. The challenge in CMA with HC is that we need to perform three inferential tasks simultaneously: approximating HC and estimating direct and indirect effects. Following the recent success of deep learning in causal inference (e.g., \cite{shalit2017estimating,louizos2017causal,swaminathan2015counterfactual}), here, we leverage deep latent-variable models that follow the causal structure of inference with proxies (Fig. \ref{graphs}) to simultaneously uncover HC and infer how it affects treatment, mediator, and outcome. Our main contributions are:
\begin{itemize}[leftmargin=*]
    \item We study a practical problem of inferring direct and indirect effects to enhance the interpretability of standard effect estimation. We circumvent the strong sequential ignorability assumptions.
    \item We extend the proxy strategy in standard causal effect estimation to CMA. We then follow the causal structure with proxies to design a deep latent-variable model that can simultaneously estimate HC and the direct and indirect effects. 
    \item Empirical evaluations on synthetic and semi-synthetic datasets show that our method outperforms existing CMA methods. We also demonstrate its applications in causal fairness analysis. 
\end{itemize}
% The remainder of this paper is organized as follows: we review the related work in Sec. 2 and describe causal effect estimation in CMA in Sec. 3. In Sec. 4, we introduce the required assumptions for identification of causal effects and the details of the proposed framework. We then present the results of empirical evaluations in Sec. 5. We summarize our findings and discuss directions for future work in Sec. 6.
\section{Related Work}
% In this section, we review common approaches to CMA as well as applications of proxy variables in causal inference literature.
\noindent\textbf{Causal Mediation Analysis.} A common approach in CMA is Linear Structural Equation Models (LSEM) \cite{baron1986moderator,mackinnon2012introduction,mackinnon1993estimating,hyman1955survey,judd1981process,james1982causal}. Under the assumption of sequential ignorability, it estimates direct and indirect effects by a system of linear equations where the mediator is a function of the treatment and the covariates, and the outcome is a function of the mediator, treatment and the covariates. The coefficients of treatment and mediator are the corresponding direct and indirect effects, respectively. Notwithstanding its appealing simplicity, the additional assumptions 
% of 1) no treatment-mediator interaction on the outcome, 2) correct parametric models, and 3) a linear relationship between the mediator and outcome 
required are often impractical, resulting in potentially inaccurate variance estimation \cite{huber2016finite,rudolph2019causal}. \citet{imai2010general} generalized LSEM by introducing a nonlinear term denoted by the interaction between the treatment and mediator. 

To address the issues, another line of research is built on the targeted maximum likelihood framework \cite{van2006targeted}. For instance,  G-computation in \cite{zheng2012targeted} allows for the treatment-mediator interactions and can handle nonlinear mediator and outcome models. Specifically, it applied the targeted maximum likelihood framework to construct ``semiparametric efficient, multiply robust, substitution estimator'' for the natural direct effect, the direct effect pertaining to an experiment where the mediator is set to null \cite{zheng2012targeted}. Another semiparametirc methodology based approach \cite{tchetgen2012semiparametric} used a general framework for obtaining inferences about natural direct and indirect effects, while accounting for pre-exposure confounding factors for the treatment and mediator. There is also work developed without the assumption of no post-treatment confounders, see e.g., \cite{rudolph2017robust,rudolph2019causal}. Huber. et al \cite{huber2013performance} employed inverse probability weighting (IPW) when estimating the direct and indirect effects to improve model's flexibility. Natural effect models are conditional mean models for counterfactuals in CMA. It relies on marginal structural models to directly parameterize the direct and indirect effects \cite{lange2012simple}. However, this simplicity comes at the price of relying on correct models for the distribution of mediator and some loss of precision.

\noindent\textbf{Proxy Variables.} Proxy variables have been widely studied in the causal inference literature \cite{nelson1991conditional,wooldridge2009estimating,louizos2017causal,miao2018identifying} given its importance to control for unobservables in observational studies. For example, Woodldridge \cite{wooldridge2009estimating} estimated firm-level production functions using proxy variables to control for the unobserved productivity. However, there have been many considerations of how to use proxies correctly \cite{wickens1972note,frost1979proxy}. McCallum \cite{mccallum1972relative} and Wickens \cite{wickens1972note} showed that the bias induced by the observables is always smaller when the proxy variable is included if the resulting measurement error is a random variable independent of the true independent variables. More recent results from \cite{frost1979proxy}, nevertheless, prove that the aforementioned conclusion, while correct, is potentially misleading. Discussions about the conditions to use proxy variables for causal identifiability can be found in \cite{cai2012identifying,greenland2011bias}. The basic idea is to infer the joint distribution of HC and the proxies, and then adjust for HC by using additional knowledge in this joint distribution \cite{wooldridge2009estimating,pearl2012measurement,kolenikov2009socioeconomic}.

More recently, Miao et al. \cite{miao2018identifying} proposed conditions under which to identify more general and complicated models with proxy variables. Specifically, the authors proved that, with at least two independent proxy variables satisfying a certain rank condition, the causal effect can be nonparametrically identified \cite{miao2018identifying}. However, in reality, we cannot know the nature of HC. Therefore, researchers have recourse to recent advances in deep learning models. Louizos et al. \cite{louizos2017causal} proposed a deep latent-variable model -- CEVAE (causal effect variational auto-encoder) -- that leverages auto-encoder to recover HC from proxies. Our work shares a similar idea of using proxies to approximate HC in the latent space. This provides us alternatives to confront the cases when \textit{sequential ignorability} assumption is violated. Nevertheless, compared to total effect estimation in \cite{louizos2017causal}, CMA needs to further distinguish and identify direct and indirect effects in different causal pathways, and account for confounders that exist before and after treatment.

Overall, previous literature in CMA relies on the stringent sequential ignorability assumptions that are often unverifiable, and generally violated in reality. Fortunately, in practice, we often observe and measure covariates that can at least partially reflect the nature of HC, i.e., proxies. In this work, we have recourse to proxy variables and the latent-variable models to approximately recover HC using observational data. The goal is to achieve more consistent and accurate estimation results.
\section{Causal Mediation Analysis}
Here, we introduce basic concepts in CMA \cite{imai2010general} under the Potential Outcome framework \cite{rubin2005causal}.
\subsection{Causal Effects in CMA} 
Let $T_i\in\{0,1\}$ be the binary treatment indicator, which is 1 if a unit $i \in \{1,2,...,n\}$ is treated, and 0 otherwise. $\mat{M}_i(t), t\in\{0,1\}$ is the mediator under treatment $t$.
We first define the conditional indirect (mediation) effect (CME) of $T$ on $Y$ via $\mat{M}(t)$ for unit $i$: 
\begin{equation}
    \delta_i(t)=Y_i(t,\mat{M}_i(1)) - Y_i(t,\mat{M}_i(0)), \qquad t=0,1,
    \label{cme}
\end{equation}
where $Y_i(\cdot)$ is the potential outcome depending on the mediator and the treatment assignment. For example, $Y_i(1,2)$ denotes the annual income a male employee $i$ being a mechanical engineer. Identification of causal effects in CMA is more challenging than total effect estimation because we can only observe partial results for both outcome and mediator, i.e., the \textit{factuals}. Suppose that $i$ is assigned to treatment $t$, we can observe $Y_i(t,\mat{M}_i(t))$ but not $Y_i(1-t,\mat{M}_i(1-t))$, $Y_i(t, \mat{M}_i(1-t))$ and $Y_i(1-t, \mat{M}_i(t))$, i.e., the \textit{counterfactuals}. Accordingly, we can define the conditional direct effect (CDE) of the treatment for each unit $i$ as follows:
\begin{equation}
    \zeta_i(t)=Y_i(1,\mat{M}_i(t))-Y_i(0,\mat{M}_i(t)), \quad t=0,1.
\end{equation}
For instance, $\zeta_i(1)$ describes the direct effect of being a male on employee $i$'s income while fixing his occupation that would be realized by being a male. 
$i$'s total effect of the treatment (e.g., individual treatment effect) is defined as the sum of direct and indirect effects:
\begin{equation}
    \tau_i=Y_i(1,\mat{M}_i(1))-Y_i(0,\mat{M}_i(0))=\frac{1}{2}\sum_{t=0}^1\{\delta_i(t)+\zeta_i(t)\}.
\end{equation}
We are typically interested in average causal mediation effect (ACME) and average causal direct effect (ACDE):
\begin{equation}
\begin{split}
    \overline{\delta}(t)=\mathbb{E}[Y_i(t,\mat{M}_i(1))-Y_i(t,\mat{M}_i(0))] \quad t=0,1;\\
    \overline{\zeta}(t)=\mathbb{E}[Y_i(1,\mat{M}_i(t))-Y_i(0,\mat{M}_i(t))] \quad t=0,1,
\end{split}
\label{ACME}
\end{equation}
where the expectations are taken over the population. $\overline{\zeta}(0)$ is also referred to as the \textit{Natural Direct Effect} \cite{pearl2014interpretation} which measures the expected increase in $Y$ as the treatment changes from $T=0$ to $T=1$ while setting mediator variable under $T=0$. Therefore, $\overline{\delta}(1)=\overline{\tau}-\overline{\zeta}(0)$ quantifies the extent to which the outcome of $Y$ is \textit{owed to} mediation \cite{pearl2014interpretation}. For simplicity and compatibility with prior benchmarks, we assume that treatment $T$ is binary and our focus in this work is on $\overline{\delta}(1)$ and $\overline{\zeta}(0)$.
% Our method can also be applied to multiple treatments or treatment with continuous values.
% \begin{table}
% \small
% \begin{center}
% \caption{Primary symbols.}
% \begin{tabular}{|c|c| } \hline
% Notation&Definition/Description\\\hline
% $\mat{X},Y$& Covariates, Outcome\\\hline
% $\mat{Z},\mat{M},T$& Hidden confounders, Mediator, Treatment\\\hline
% $G$&Causal graph in Fig. \ref{graphs}\\\hline
% $D_x, D_z$& The dimension of $\mat{X}$ and $\mat{Z}$\\\hline
% $n$& Sample size\\\hline
% $\delta_i(t)$& Indirect effect of unit $i$ under treatment $t$\\\hline
% $\zeta_i(t)$& Direct effect of unit $i$ under treatment $t$\\\hline
% $\tau_i$ & Total effect or individual treatment effect\\\hline
% $f_k,g_k$& A neural network\\ \hline
% $\bm{\theta}_k,\bm{\phi}_k$& Parameters of neural network $f_k,g_k$\\\hline
% $t^*_i$ &Observed value for the treatment of unit $i$\\ \hline
% $do(t=1)$& Intervene on the treatment\\\hline
% ACME($\overline{\delta}(t)$)& Average causal mediation effect under treatment $t$\\\hline
% ACDE($\overline{\zeta}(t)$)&Average causal direct effect under treatment $t$\\\hline
% ATE($\overline{\tau}$)&Average treatment effect\\\hline
% \end{tabular}
% \label{notations}
% \end{center}
% \end{table}
\subsection{Sequential Ignorability Assumption} 
\begin{assumption}[Sequential Ignorability \cite{imai2010general}] 
\begin{gather}
    {Y_i(t',\mat{m}),\mat{M}_i(t)}\indep T_i|\mat{X}_i=\mat{x},\\
     Y_i(t',\mat{m})\indep \mat{M}_i(t)| T_i=t,\mat{X}_i=\mat{x},
\end{gather}
where  $\mat{x}\in \mat{X}$ and $\mat{m} \in \mat{M}$, $0<Pr(T_i=t|\mat{X}_i=\mat{x})<1$, $0<Pr(\mat{M}_i(t)=\mat{m}|T_i=t,\mat{X}_i=\mat{x})<1$. $t, t'\in\{0,1\}$.
\end{assumption}
The first ignorability assumption is identical to the strong ignorability in estimating the total effect, or average treatment effect (ATE) \cite{pearl2010consistency,keele2015causal}. It assumes that the treatment assignment is ignorable, i.e., statistically independent of potential outcomes and potential mediators given covariates $\mat{X}$. The second ignorability describes that mediator is independent of outcome conditional on treatment and covariates. The sequential ignorability implies that the same set of covariates $\mat{X}$ can account for the confounding bias in both the treatment- and mediator-outcome relationships. This is a strong assumption because typically we cannot rule out the possibility of \textit{hidden confounding bias} in observational studies.
\section{Estimating Effects in CMA}
The goal is to circumvent the stringent sequential ignorability assumption to make CMA more applicable.
We therefore consider HC, which we assume can be inferred in the latent space through proxy variables that are closely related to HC. Without knowing the exact nature of HC, we leverage recent advances in deep latent-variable models that closely follow the causal graph in Fig. \ref{graphs}. The proposed model can simultaneously uncover HC and infer how HC affects treatment, mediator, and outcome. 
\begin{figure*}
    \centering
      \includegraphics[width=.75\linewidth]{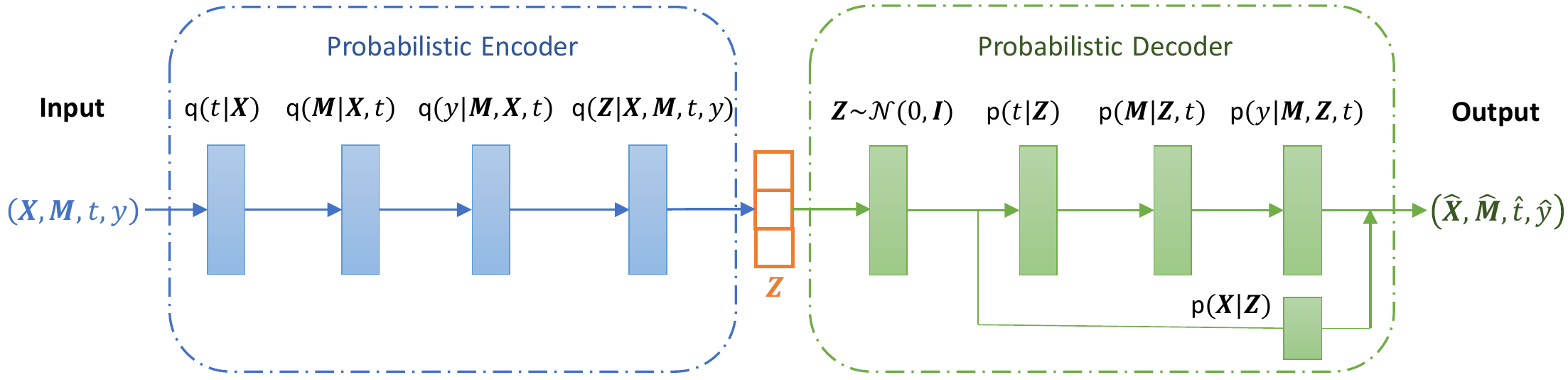}
    \caption{Illustration of the overall architecture of the networks for CMAVAE. The Probabilistic Encoder (blue part) takes the observables $(\mat{X},\mat{M},t,y)$ as the input and encode the posterior distribution $q(\mat{Z}|\mat{X},\mat{M},t,y)$. The Probabilistic Decoder (green part) takes the input of $\mat{Z}$ sampled from Gaussian prior distribution and decode the prior distributions for $(\mat{X},\mat{M},t,y)$. Best viewed in colors.}
\label{VAE}
\end{figure*}
\subsection{Identifying Causal Mediation Effect}
To recover ACME and ACDE from observational data, we take the expectation of CME and CDE under treatment $t$. First, we introduce the following assumption that validates the identification results:
\begin{assumption}[The Proposed Assumption]\ \\
\begin{itemize}
    \item There exists some latent variable $\mat{Z}$ that simultaneously deconfounds $T, \mat{M}$, and $Y$. Formally,
    \begin{gather}
    {Y(t',\mat{m}),\mat{M}(t)}\indep T|\mat{Z}=\mat{z},\\
     Y(t',\mat{m})\indep \mat{M}(t)| T=t,\mat{Z}=\mat{z}.
\end{gather}
    \item  There exists some observed variable $\mat{X}$ that approximates $\mat{Z}$;
    \item $p(\mat{Z},\mat{X},\mat{M},t,y)$ can be approximately recovered from the observations $(\mat{X},\mat{M},t,y)$ under Fig. \ref{graphs}. 
\end{itemize}
\end{assumption}
\noindent The proposed assumption ensures that the HC $\mat{Z}$ can be (at least partially) estimated by covariates $\mat{X}$, $T$, $\mat{M}$ and $Y$. While HC is not necessarily always related to $\mat{X}$, there are many cases where this is possible \cite{louizos2017causal}.
Similar to previous literature \cite{cai2012identifying} in causal inference in the presence of HC, the third assumption also implies that the causal graph $G$ and the corresponding joint distribution $p(\mat{Z},\mat{X},\mat{M},t,y)$ are faithful to each other, i.e., the conditional independence in the joint distribution is also reflected in $G$, and vice versa \cite{spirtes2000causation}. 

Given observations ($\mat{X},\mat{M},t,y$), ACME in Eq. (\ref{ACME}) can then be reformulated as 
\begin{equation}
\begin{split}
    \overline{\delta}(t)&:=\mathbb{E} [CME(\mat{x},t)], \text{with}\\
     &CME(\mat{x},t):=\mathbb{E}[y|\mat{X}=\mat{x}, T=t, \mat{M}(do(t'=1))]\\
     &-\mathbb{E}[y|\mat{X}=\mat{x}, T=t,  \mat{M}(do(t'=0))], \quad t=0,1.
\end{split}
\end{equation}
Based on previous results in \cite{louizos2017causal} and Pearl's back-door adjustment formula \cite{pearl2012measurement}, we propose the following theorem:
\begin{theorem*}
% Assuming that the value of covariates is unique for each unit, 
If we estimate $p(\mat{Z},\mat{X},\mat{M},t,y)$, then we recover CME and CDE under the causal graph in Fig. \ref{graphs}.
\end{theorem*}

The proof can be seen in Appendix A. Theorem 1 connects the joint distribution $p(\mat{Z},\mat{X},\mat{M},t,y)$ with causal effects in CMA. We recognize that the true distribution of $p(\mat{Z},\mat{X},\mat{M},t,y)$ can only be approximately recovered using observational data. As noted in the Related Work section, literature in various research fields also provides support for using the joint distribution to study causal effects with HC. We do acknowledge the weakness of the approximated $p(\mat{Z},\mat{X},\mat{M},t,y)$ in making conclusive causal claims, but they provide complementary advantages over conventional sequential ignorability assumptions in many aspects. %
Particularly noteworthy is that the observed covariates $\mat{X}$ should be related to $\mat{Z}$. For example, $\mat{X}$ is a ``noisy'' function of $\mat{Z}$ which is comprised of binary variables \cite{jernite2013discovering}. Next, we show how to approximate $p(\mat{Z},\mat{X},\mat{M},t,y)$ from observations of $(\mat{X},\mat{M},t,y)$ using deep latent variable models. 
\subsection{Causal Mediation Analysis Variational Auto-Encoder}
Our approach--\textbf{C}ausal \textbf{M}ediation \textbf{A}nalysis with \textbf{V}ariational \textbf{A}uto-\textbf{E}ncoder (CMAVAE)--builds on Variational Auto-Encoder (VAE) \cite{kingma2013auto} that discovers $\mat{Z}$ in the latent space by variational inference. VAE makes smaller error in approximations and is robust against the noise of proxy variables. Here, we use VAEs to infer the complex non-linear relationships between $\mat{X}$ and $(\mat{Z},\mat{M},t,y)$, and approximately recover $p(\mat{Z},\mat{X},\mat{M},t,y)$. CMAVAE parameterizes the causal graph in Fig. \ref{graphs} as a latent-variable model with neural network functions connecting the variables of interest. The objective function of VAE is then the reconstruction error of the observed $(\mat{X},\mat{M},t,y)$ and the inferred $(\hat{\mat{X}},\hat{\mat{M}},\hat{t},\hat{y})$. Fig. \ref{VAE}. features the overall architecture design of CMAVAE. In the following descriptions of VAE, $\mat{x}_i$ denotes a feature vector of an input sample $i$, $\mat{m}_i$ is the mediator, $t_i$, $y_i$, and $\mat{z}_i$ denote the treatment status, outcome, and HC, respectively.

Let $D_x$, $D_z$ be the dimensions of $\mat{x}_i$ and of $\mat{z}_i$ respectively and each $f_k(\cdot), k\in\{1,2,3,4,5\}$ represents a neural network parameterized by its own parameters $\bm{\theta}_k$. We define
\begin{gather}
    p(\mat{z}_i)=\prod_{j=1}^{D_z}\mathcal{N}(\mat{z}_{ij}|0,1); \quad p(\mat{x}_i|\mat{z}_i)=\prod_{j=1}^{D_x}p(\mat{x}_{ij}|\mat{z}_i); \\ p(t_i|\mat{z}_i)=\text{Bern}(\sigma(f_1(\mat{z}_i))),
\end{gather}
with $p(\mat{x}_{ij}|\mat{z}_i)$ being a probability distribution for the $j$-th covariate of $\mat{x}_i$ and $\sigma(\cdot)$ being the logistic function. For a continuous mediator and outcome\footnote{Please refer to Appendix B for binary cases.}, we parameterize the probability distribution as Gaussian distribution. The corresponding parameters are designed by using a similar architecture\footnote{We use simple concatenation rather than the shared representations as in TARnet because preliminary experiments show that they achieve similar performances.} inspired by TARnet \cite{shalit2017estimating}.
The details can be seen as follows:
\begin{gather}
    p(\mat{m}_i|\mat{z}_i,t_i)=\mathcal{N}(\mu=\hat{\mu}_{1i},\sigma^2=\hat{\nu}_1);\\
     p(y_i|\mat{m}_i,\mat{z}_i,t_i)=\mathcal{N}(\mu=\hat{\mu}_{2i},\sigma^2=\hat{\nu}_2),
    %p(\mat{m}_i|\mat{z}_i,t_i)=\text{Bern}(\pi=\hat{\pi}_{1i});\\
    % p(y_i|\mat{m}_i,\mat{z}_i,t_i)= \text{Bern}(\pi=\hat{\pi}_{2i}),
\end{gather}
\begin{gather}
    \hat{\mu}_{1i}=t_if_2(\mat{z}_i)+(1-t_i)f_3(\mat{z}_i); \\
    \hat{\mu}_{2i}=t_if_4(\mat{z}_i\circ \mat{m}_i)+(1-t_i)f_5(\mat{z}_i\circ \mat{m}_i).
    %\hat{\pi}_{1i}=\sigma(t_if_2(\mat{z}_i)+(1-t_i)f_3(\mat{z}_i));\\
    %\hat{\pi}_{2i}=\sigma(t_if_4(\mat{z}_i\circ \mat{m}_i)+(1-t_i)f_5(\mat{z}_i\circ \mat{m}_i)).
\end{gather}
$\hat{\nu}_1$ and $\hat{\nu}_2$ are predefined constant. We denote the concatenation of vectors as $\circ$. Then the posterior over $\mat{Z}$ can be approximated by 
\begin{gather}
    q(\mat{z}_i|\mat{x}_i,\mat{m}_i,y_i,t_i)=\prod_{j=1}^{D_z}\mathcal{N}(\mu_j=\overline{\mu}_{ij},\sigma^2_j=\overline{\sigma}^2_{ij});\\
    \overline{\bm{\mu}}_i=t_i\bm{\mu}_{t=0,i}+(1-t_i)\bm{\mu}_{t=1,i};\\ \overline{\bm{\sigma}}_i^2=t_i\bm{\sigma}^2_{t=0,i}+(1-t_i)\bm{\sigma}^2_{t=1,i};\\
    \bm{\mu}_{t=0,i},\bm{\sigma}^2_{t=0,i}=g_1(\mat{x}_i\circ y_i \circ \mat{m}_i);\\
     \bm{\mu}_{t=1,i},\bm{\sigma}^2_{t=1,i}=g_2(\mat{x}_i\circ y_i \circ \mat{m}_i), 
\end{gather}
where $g_k$ is a neural network with variational parameters $\bm{\phi}_k$ for $k\in\{1,2,...,7\}$. The objective function of VAEs is then:
\begin{gather}
\begin{split}
    \mathcal{L}=\sum_{i=1}^n\mathbb{E}_{q(\mat{z}_i|\mat{x}_i,t_i,\mat{m}_i,y_i)}[\log p(\mat{z}_i)+\log p(\mat{x}_i|\mat{z}_i)+\log p(t_i|\mat{z}_i)\\
    +\log p(\mat{m}_i|\mat{z}_i,t_i)
    +\log p(y_i|\mat{m}_i,\mat{z}_i,t_i)-\log q(\mat{z}_i|\mat{x}_i,\mat{m}_i,y_i,t_i)].
\end{split}
\end{gather}
However, in the Encoder, we still need to infer $t$, $\mat{m}$ and $y$ for new subjects before inferring the posterior distribution over $\mat{z}$. Hence, we introduce three auxiliary distributions that predict $t_i, \mat{m}_i, y_i$ for new sample characterized by $\mat{x}_i$:
\begin{gather}
 q(t_i|\mat{x}_i)=\text{Bern}(\pi=\sigma(g_3(\mat{x}_i)));\\
 q(\mat{m}_i|\mat{x}_i,t_i)=\mathcal{N}(\mu=\overline{\mu}_{1i},\sigma^2=\overline{\nu}_1);\\
    \overline{\mu}_{1i}=t_ig_4(\mat{x}_i)+(1-t_i)g_5(\mat{x}_i).
   % \overline{\pi}_{1i}=\sigma(t_ig_4(\mat{x}_i)+(1-t_i)g_5(\mat{x}_i)).
\end{gather}
Similarly, we have 
\begin{gather}
q(y_i|\mat{x}_i,\mat{m}_i,t_i)=\mathcal{N}(\mu=\overline{\mu}_{2i},\sigma^2=\overline{\nu}_2);\\
    \overline{\mu}_{2i}=t_ig_{6}(\mat{x}_i\circ \mat{m}_i)+(1-t_i)g_7(\mat{x}_i\circ \mat{m}_i),
    %\overline{\pi}_{2i}=\sigma(t_ig_{6}(\mat{x}_i\circ \mat{m}_i)+(1-t_i)g_7(\mat{x}_i\circ \mat{m}_i)).
\end{gather}
Here, $\overline{\nu}_1$ and $\overline{\nu}_2$ are predefined. 
% We add these auxiliary terms to the objective function in order to ``learn'' the parameters of $q(t|\mat{x})$, $q(\mat{m}|\mat{x},t)$ and $q(y|\mat{x},\mat{m},t)$.
The final loss is defined as 
\begin{equation}
    \begin{split}
        \mathcal{F}&=\mathcal{L}+\sum_{i=1}^n (\log q(\mat{m}_i=\mat{m}_i^*|\mat{x}_i^*,t_i^*)+\\
        \log q(t_i&=t_i^*|\mat{x}_i^*)+\log q(y_i=y_i^*|\mat{x}_i^*,\mat{m}_i^*,t_i^*)),
    \end{split}
\end{equation}
where $y_i^*,\mat{x}_i^*,\mat{m}_i^*,t_i^*$ are the observed values for the outcome, input, mediator, and treatment random variables in the training set. 
% To infer ACME under treated ($\overline{\delta}(1)$), we first randomly sample $\mat{Z}$ and denote the results as $\hat{\mat{Z}}$. We then obtain the estimations of $\mat{M}(1)$ and $\mat{M}(0)$ by $\hat{\mat{m}}_1\sim p(\mat{m}|\mat{z}=\hat{\mat{z}},t=1)$ and $\hat{\mat{m}}_0\sim p(\mat{m}|\mat{z}=\hat{\mat{z}},t=0)$,  respectively. Similarly, we estimate outcome $Y(1,M(1))$ by $\hat{\mat{y}}_{1,m_1}\sim p(\mat{y}|\mat{z}=\hat{\mat{z}},\mat{m}=\hat{\mat{m}}_1,t=1)$ and $Y(1,M(0))$ by $\hat{\mat{y}}_{1,m_0}\sim p(\mat{y}|\mat{z}=\hat{\mat{z}},\mat{m}=\hat{\mat{m}}_0,t=1)$. Lastly, we take the average of $\hat{\mat{y}}_{1,m_1}-\hat{\mat{y}}_{1,m_0}$ to get the estimation of $\overline{\delta}(1)$. We use a similar procedure to get the inferred ACDE under control $\overline{\zeta}(0)$.
%
%
\section{Experiments}
Evaluation in causal inference has been a long-standing challenge due to the lack of ground-truth effects \cite{guo2020survey,cheng2022evaluation,cheng2019practical}. For CMA, this is more challenging because both direct and indirect effects need to be specified, which cannot be achieved by randomized experiments.
Therefore, CMA evaluation often has to rely on simulations \cite{huber2016finite,imai2013experimental}. Here, we first conduct empirical evaluations using synthetic and semi-synthetic datasets, where real data is modified such that the true indirect and direct effects are known. We also perform a case study to show the application of CMAVAE in fair machine learning. 
\subsection{Experimental Settings}
We used Tensorflow \cite{abadi2016tensorflow} and Edward \cite{tran2016edward} to implement CMAVAE. Design of the neural network architecture is similar to that in \cite{shalit2017estimating}. For JOBS II data, unless otherwise specified, we used 5 hidden layers with size 100 and with ELU \cite{clevert2015fast} nonlinearities for the approximate posterior over the latent variables $q(\mat{Z}|\mat{X},\mat{M},t,y)$, the generative model $p(\mat{X}|\mat{Z})$, the mediator models $p(\mat{M}|\mat{Z},t)$, $q(\mat{M}|\mat{X},t)$ and the outcome models $p(Y|\mat{M},\mat{Z},t)$, $q(Y|\mat{M},\mat{X},t)$. One single hidden layer neural network with ELU nonlinearities is adopted to model $p(t|\mat{Z})$, $q(t|\mat{X})$. 
The dimension of the latent variable $\mat{Z}$ is set to 10. To prevent overfitting in the neural nets, we also applied a small weight decay term to all parameters with $\lambda=10^{-3}$. We optimized the objective function with Adam \cite{kingma2014adam} and a learning rate of $10^{-6}$. To estimate the mediators and outcomes, we averaged over 100 samples from the approximate posterior $q(\mat{Z}|\mat{X})=\sum_t\int_y q(\mat{Z}|t,y,\mat{X},\mat{M})q(y|t,\mat{X},\mat{M})q(\mat{M}|t,\mat{X})q(t|\mat{X})dy$. More implementation details can be found in Appendix C.

Baselines include popular parametric methods LSEM \cite{baron1986moderator}, LSEM-I \cite{imai2010general}, NEM-W, NEM-I \cite{lange2012simple}, and semi-parametric methods IPW \cite{huber2014identifying}:
\begin{itemize}[leftmargin=*]
    \item \textbf{LSEM} \cite{baron1986moderator}. A common framework for CMA. It is based on linear equations characterizing the outcome and the mediator.
    \item \textbf{LSEM-I} \cite{imai2010general}. An non-linear extension of LSEM with an interaction term between treatment and mediator. 
    \item \textbf{IPW} \cite{huber2014identifying,huber2013performance}. This approach is based on a common weighting strategy in causal inference, i.e.,IPW \cite{robins1994estimation}, to measure confounder effects and adjust analyses to remove the confounding bias.
    \item \textbf{Natural Effect Models (NEM)} \cite{lange2012simple}. NEM are conditional mean models for counterfactuals in CMA. It directly models the direct and indirect effects by treating the task as a missing data problem. There are two approaches for data augmentation in NEM according to unobserved $(t, 1-t)$ combinations -- the weighting-based approach (\textbf{NEM-W}) and the imputing-based approach (\textbf{NEM-I}).
\end{itemize}
We used R package ``mediation''\footnote{https://cran.r-project.org/web/packages/mediation/index.html} for LSEM and LSEM-I, ``causalweight''\footnote{https://cran.r-project.org/web/packages/causalweight/index.html} for IPW (the parameter \textit{trim} was set to 0.05), and ``medflex''\footnote{https://cran.r-project.org/package=medflex} for NEM-W, NEM-I. Evaluation metric is the absolute error of estimated ACDE $\overline{\zeta}(0)$, ACME $\overline{\delta}(1)$, and the total effect (ATE). 
\subsection{JOBS II Dataset}
We first test CMAVAE on the real-world dataset JOBS II \cite{vinokur1999jobs}, collected from a randomized field experiment that investigates the efficacy of a job training intervention on unemployed workers. Participants were randomly asked to  attend the job skills workshops, i.e., the treatment group, or receive a booklet, i.e., the control group. The treatment group learned job hunting skills and coping strategies for confronting the setbacks in the job hunting process while the control group learned job hunting tips through a booklet. In follow-up interviews, the outcome -- a continuous measure of depressive symptoms -- was measured. Mediator $M$ is a continuous measure representing the job search self-efficacy. JOB II can be downloaded from R package ``mediation'' and the total sample size is 899. We further performed normalization on all continuous covariates and applied one-hot encoding to categorical covariates. 
% These covariates include education level, income, age, race, sex, marital status, previous occupation, and the level of economic hardship.
\begin{table*}[ht!]
\setlength\tabcolsep{2pt}
\begin{center}
\caption{Absolute errors for JOBS II data with 10\% mediated ($\alpha$). $n=500,1000$ and $\eta=1, 10$. (\%)}
\begin{tabular}{ labababababab } \hline
\rowcolor{white}
\multicolumn{13}{c}{\textbf{ACME under treated} ($\overline{\delta}(1)$)}\\\hline
\rowcolor{white}
Models&\multicolumn{2}{c}{LSEM}&\multicolumn{2}{c}{LSEM-I}&\multicolumn{2}{c}{NEM-W}&\multicolumn{2}{c}{NEM-I}&\multicolumn{2}{c}{IPW}&\multicolumn{2}{c}{CMAVAE}\\\cmidrule(lr){2-3} \cmidrule(lr){4-5}\cmidrule(lr){6-7} \cmidrule(lr){8-9}\cmidrule(lr){10-11} \cmidrule(lr){12-13}
$n$&500&1000&500&1000&500&1000&500&1000&500&1000&500&1000\\\hline
$\eta=10$&0.7$\pm$.03&0.7$\pm$.02&0.9$\pm$.04&0.6$\pm$.02&0.3$\pm$.04&0.3$\pm$.01&0.6$\pm$.03&0.8$\pm$.01&0.6$\pm$.04&0.8$\pm$.02&\textbf{0.2$\pm$.00}&\textbf{0.3$\pm$.00}\\
$\eta=1$&0.1$\pm$.01&0.1$\pm$.01&0.0$\pm$.01&0.1$\pm$.01&0.0$\pm$.01&0.1$\pm$.01&\textbf{0.0$\pm$.00}&0.1$\pm$.01&0.0$\pm$.01&0.1$\pm$.01&0.1$\pm$.00&\textbf{0.1$\pm$.00}\\
\multicolumn{13}{c}{\textbf{ACDE under control} ($\overline{\zeta}(0)$)}\\\hline
$\eta=10$&0.8$\pm$.07&2.0$\pm$.06&1.3$\pm$.07&1.6$\pm$.06&2.5$\pm$.06&1.2$\pm$.05&1.2$\pm$.06&1.8$\pm$.05&1.2$\pm$.06&0.2$\pm$.06&\textbf{0.1$\pm$.00}&\textbf{0.0$\pm$.03}\\
$\eta=1$&3.2$\pm$.08&0.3$\pm$.07&3.3$\pm$.08&\textbf{0.0$\pm$.07}&3.3$\pm$.08&0.3$\pm$.07&1.1$\pm$.03&0.2$\pm$.07&3.3$\pm$.08&0.3$\pm$.06&\textbf{0.5$\pm$.02}&0.4$\pm$.01\\
\multicolumn{13}{c}{\textbf{ATE} ($\overline{\tau}$)}\\\hline
$\eta=10$&1.5$\pm$.06&1.2$\pm$.05&2.2$\pm$.05&1.0$\pm$.06&2.2$\pm$.06&0.8$\pm$.06&1.8$\pm$.05&0.9$\pm$.06&0.5$\pm$.05&1.0$\pm$.06&\textbf{0.3$\pm$.01}&\textbf{0.3$\pm$.03}\\
$\eta=1$&3.4$\pm$.08&0.2$\pm$.07&3.3$\pm$.08&0.1$\pm$.07&3.4$\pm$.08&0.2$\pm$.06&3.4$\pm$.03&\textbf{0.1$\pm$.06}&3.2$\pm$.07&0.2$\pm$.05&\textbf{0.4$\pm$.02}&0.3$\pm$.01\\\hline
\end{tabular}
\label{jobs1}
\end{center}
\end{table*}
\begin{table*}[ht!]
\setlength\tabcolsep{2pt}
\begin{center}
\caption{Absolute errors for JOBS II data with 50\% mediated ($\alpha$). $n=500,1000$ and $\eta=1, 10$. (\%)}
\begin{tabular}{ labababababab } \hline
\rowcolor{white}
\multicolumn{13}{c}{\textbf{ACME under treated} ($\overline{\delta}(1)$)}\\\hline
\rowcolor{white}
Models&\multicolumn{2}{c}{LSEM}&\multicolumn{2}{c}{LSEM-I}&\multicolumn{2}{c}{NEM-W}&\multicolumn{2}{c}{NEM-I}&\multicolumn{2}{c}{IPW}&\multicolumn{2}{c}{CMAVAE}\\\cmidrule(lr){2-3} \cmidrule(lr){4-5}\cmidrule(lr){6-7} \cmidrule(lr){8-9}\cmidrule(lr){10-11} \cmidrule(lr){12-13}
$n$&500&1000&500&1000&500&1000&500&1000&500&1000&500&1000\\\hline
$\eta=10$&1.3$\pm$.03&0.5$\pm$.03&0.9$\pm$.03&0.6$\pm$.03&0.9$\pm$.04&1.2$\pm$.03&0.2$\pm$.03&0.4$\pm$.03&0.2$\pm$.03&0.4$\pm$.03&\textbf{0.0$\pm$.00}&\textbf{0.1$\pm$.00}\\
$\eta=1$&0.2$\pm$.01&\textbf{0.0$\pm$.01}&0.1$\pm$.01&\textbf{0.0$\pm$.01}&0.2$\pm$.00&0.1$\pm$.00&0.2$\pm$.00&0.1$\pm$.01&0.1$\pm$.01&\textbf{0.0$\pm$.01}&\textbf{0.1$\pm$.00}&0.1$\pm$.00\\
\multicolumn{13}{c}{\textbf{ACDE under control} ($\overline{\zeta}(0)$)}\\\hline
$\eta=10$&0.9$\pm$.07&0.1$\pm$.04&0.6$\pm$.06&0.1$\pm$.04&0.2$\pm$.07&0.5$\pm$.04&\textbf{0.1$\pm$.06}&0.1$\pm$.04&0.7$\pm$.07&0.2$\pm$.05&0.3$\pm$.01&\textbf{0.1$\pm$.00}\\
$\eta=1$&0.4$\pm$.01&0.4$\pm$.01&0.1$\pm$.10&0.3$\pm$.10&0.5$\pm$.10&0.4$\pm$.04&0.1$\pm$.10&0.3$\pm$.04&0.3$\pm$.10&0.2$\pm$.04&\textbf{0.1$\pm$.00}&\textbf{0.1$\pm$.00}\\
\multicolumn{13}{c}{\textbf{ATE} ($\overline{\tau}$)}\\\hline
$\eta=10$&0.4$\pm$.05&0.6$\pm$.03&0.3$\pm$.05&0.8$\pm$.03&0.7$\pm$.05&0.7$\pm$.01&\textbf{0.1$\pm$.05}&0.5$\pm$.03&0.9$\pm$.05&0.2$\pm$.04&0.3$\pm$.01&\textbf{0.0$\pm$.01}\\
$\eta=1$&0.2$\pm$.10&0.4$\pm$.04&0.1$\pm$.09&0.3$\pm$.04&0.2$\pm$.10&0.3$\pm$.04&0.3$\pm$.10&0.3$\pm$.04&0.2$\pm$.10&0.2$\pm$.04&\textbf{0.0$\pm$.01}&\textbf{0.2$\pm$.01}\\\hline
\end{tabular}
\label{jobs2}
\end{center}
\end{table*}

\noindent\textbf{Simulation.} To obtain the ground truth for direct and indirect effects, we used a simulation approach similar to \cite{huber2013performance,huber2016finite}. Specifically, we first estimate probit specifications in which we regress (1) $T$ on $\mat{X}$ and (2) $M$ on $T$ and $\mat{X}$ using the entire JOBS II data. As the mediator in JOBS II is a continuous variable in the range of [1,5], we apply the Indicator function $I$ to $M_i$: $I\{M_i\geq 3\}$. The output of $I(\cdot)$ is 1, i.e., mediated, if the argument is satisfied and 0, i.e., nonmediated, otherwise.
All observations with $T=1$ and $I(M_i)=1$ (or both) are then discarded. We draw independent Monte Carlo samples with replacement $\mat{X}'$ from the original dataset of the nontreated and nonmediated \cite{huber2016finite}. The next step simulates the (pseudo-)treatment:
\begin{equation}
    T_i=I\{\mat{X}'_i\hat{\beta}_{pop}+U_i>0\},
\end{equation}
where $\hat{\beta}_{pop}$ are the probit coefficient estimates of the treatment model using original data. $U_i\sim \mathcal{N}(0,1)$ denotes the environment noise. The (pseudo-)mediator is simulated by 
\begin{equation}
    M_i=\eta(T_i\hat{\gamma}_{pop}+\mat{X}_i'\hat{\omega}_{pop})+\alpha+V_i,
\end{equation}
where $\hat{\gamma}_{pop}$ and $\hat{\omega}_{pop}$ are the probit coefficient estimates on $T$ and $\mat{X}$ of the mediator model using original data and $V_{i}\sim \mathcal{N}(0,1)$ denotes the Gaussian noise. $\eta$ gauges the magnitude of selection into the mediator. Following \cite{huber2016finite}, we consider $\eta=1$ (normal selection) and $\eta=10$ (strong selection). $\alpha$ determines the shares of mediated individuals. In our simulations, we set $\alpha$ such that either 10\% or 50\% of the observations are mediated ($I(M_i)=1$). With this particular simulation design, we obtain the true direct, indirect and total effect of zero. This is because all observations including the pseudo-treated/mediated ones are drawn from the population neither treated nor mediated. Combinations of the sample size, strength of selection into the mediator and share of mediated observations yield all in all 8 different data generating processes (DGPs). We use 80\% of data for training and keep the ratio of number of treated to controlled in the training and test datasets same.
\begin{table}[ht!]
\begin{center}
\caption{Performance comparisons using JOBS II data with 50\% mediated ($\alpha$), $n=500$, and $\eta=1$. The data is generated with the injection of different levels of proxy noise.}
\begin{subtable}{\columnwidth}
\centering
    \caption{Results for estimating ACME ($\overline{\delta}(1)$). (\%)}
    \begin{tabular}{|c|c|c|c|c|c|}\hline
         Proxy noise&$0.1$&$0.2$&$0.3$&$0.4$&$0.5$\\ \hline\hline
         LSEM & 0.2 & \textbf{0.1} & 0.2 & 0.2 & 0.5\\ \hline
         LSEM-I & 0.2 & 0.2 & 0.3 & 0.2 & 0.8 \\ \hline
         IPW & 0.2 & 0.2 & 0.2 & 0.2 & 0.8 \\ \hline
            NEM-W & 0.2 & \textbf{0.1} & \textbf{0.1} & 0.2 & 0.5 \\ \hline
            NEM-I & 0.2 & \textbf{0.1} & 0.2 & 0.2 & 0.5 \\\hline
         CMAVAE &\textbf{0.1} & \textbf{0.1} & 0.2 & \textbf{0.2} & \textbf{0.2}\\ \hline
    \end{tabular}
\end{subtable}
\begin{subtable}{\columnwidth}
\centering
     \caption{Results for estimating ACDE ($\overline{\zeta}(0)$). (\%)}
    \begin{tabular}{|c|c|c|c|c|c|}\hline
         Proxy noise&$0.1$&$0.2$&$0.3$&$0.4$&$0.5$\\ \hline\hline
         LSEM&2.9 & 2.8 & 4.6 &5.0 & 5.9\\\hline
         LSEM-I& 3.4 & 3.1 & 5.1 & 5.2 & 6.0\\\hline
         IPW & 3.7 & 2.8 & 4.7 & 5.3 & 5.6\\\hline
         NEM-W & 3.4 & 2.8 & 4.5 & 5.2 &5.7 \\\hline
         NEM-I & 3.4 & 2.7 & 4.6 & 5.3 & 5.8 \\\hline
         CMAVAE&\textbf{1.3} & \textbf{1.0} & \textbf{1.3} & \textbf{1.5} & \textbf{3.4}\\ \hline
    \end{tabular}
\end{subtable}
\begin{subtable}{\columnwidth}
\centering
    \caption{Results for estimating ATE ($\overline{\tau}$). (\%)}
    \begin{tabular}{|c|c|c|c|c|c|}\hline
         Proxy noise&$0.1$&$0.2$&$0.3$&$0.4$&$0.5$\\ \hline\hline
         LSEM&3.0 & 2.9 & 4.6 & 5.0 & 6.0\\\hline
         LSEM-I&3.5 & 3.2 & 5.2 & 5.1 & 5.9\\\hline
         IPW&3.8 & 2.9 & 4.7 & 5.2 & 5.8 \\\hline
         NEM-W&3.5 & 2.8 & 4.5 & 5.2 & 6.0 \\\hline
         NEM-I&3.5 & 2.8 & 4.6 & 5.2 & 6.0 \\\hline
         CMAVAE&\textbf{1.3} & \textbf{1.0} & \textbf{1.3} & \textbf{1.5} & \textbf{3.6}\\ \hline
    \end{tabular}
\end{subtable}
\label{Confounding_Noise1}
\end{center}
\end{table}
\begin{table}[ht!]
\begin{center}
\caption{Performance comparisons using JOBS II data with 10\% mediated ($\alpha$), $n=500$, and $\eta=1$. The data is generated with the injection of different levels of proxy noise.}
\begin{subtable}{\columnwidth}
\centering
 \caption{Results for estimating ACME ($\overline{\delta}(1)$). (\%)}
    \begin{tabular}{|c|c|c|c|c|c|}\hline
         Proxy noise&$0.1$&$0.2$&$0.3$&$0.4$&$0.5$\\ \hline\hline
         LSEM & 0.2 & 0.2 & \textbf{0.1} & \textbf{0.1} & 0.3\\ \hline
         LSEM-I & 0.2 & 0.2 & \textbf{0.1 }& 0.2 & 0.3 \\ \hline
         IPW & 0.2 & 0.3 & \textbf{0.1} & 0.2 & 0.2 \\ \hline
            NEM-W & 0.2 & 0.2 & \textbf{0.1} & \textbf{0.1} & 0.2 \\ \hline
            NEM-I & 0.2 & 0.2 & \textbf{0.1} & \textbf{0.1} & 0.3 \\\hline
         CMEVAE &\textbf{0.1} & \textbf{0.1} & \textbf{0.1 }& \textbf{0.1} & \textbf{0.1}\\ \hline
    \end{tabular}
\end{subtable}
\begin{subtable}{\columnwidth}
\centering
 \caption{Results for estimating ACDE ($\overline{\zeta}(0)$). (\%)}
    \begin{tabular}{|c|c|c|c|c|c|}\hline
         Proxy noise&$0.1$&$0.2$&$0.3$&$0.4$&$0.5$\\ \hline\hline
         LSEM&3.4 & 3.1 & 4.0 & 4.4 & 3.6\\\hline
         LSEM-I& 3.1 & 3.6 & 4.8 & 4.0 & 3.2\\\hline
         IPW & 3.7 & 4.0 & 4.6 & 4.1 & 2.8\\\hline
         NEM-W & 3.0 & 3.4 & 4.3 & 4.4 & 3.3 \\\hline
         NEM-I & 3.0 & 3.6 & 4.3 & 4.4 & 3.3 \\\hline
         CMEVAE&\textbf{1.6} & \textbf{1.1} & \textbf{1.6} & \textbf{1.3} & \textbf{1.5}\\ \hline
    \end{tabular}
\end{subtable}
\begin{subtable}{\columnwidth}
\centering
 \caption{Results for estimating ATE ($\overline{\tau}$). (\%)}
    \begin{tabular}{|c|c|c|c|c|c|}\hline
         Proxy noise&$0.1$&$0.2$&$0.3$&$0.4$&$0.5$\\ \hline\hline
         LSEM&3.4 & 3.0 & 4.0 & 4.4 & 3.5\\\hline
         LSEM-I&3.1 & 3.5 & 4.8 & 4.0 & 3.2\\\hline
         IPW&3.7 & 3.9 & 4.5 & 4.1 & 2.9 \\\hline
         NEM-W&3.0 & 3.3 & 4.2 & 4.4 & 3.3 \\\hline
         NEM-I&3.0 & 3.4 & 4.3 & 4.4 & 3.2\\\hline
         CMEVAE&\textbf{1.7} & \textbf{1.0} & \textbf{1.7} & \textbf{1.3} & \textbf{1.6}\\ \hline
    \end{tabular}
\end{subtable}
\label{Confounding_Noise2}
\end{center}
\end{table}

\noindent\textbf{Results.} Each set of experiment was conducted 10 replications and we report the averaged results as well as the standard deviations for the 8 DGPs in Table \ref{jobs1}-\ref{jobs2}. Note when the best mean of multiple models are equal, we highlight results with the least standard deviations. We can draw following conclusions from our observations:
\begin{itemize} [leftmargin=*]
    \item CMAVAE mostly outperforms the baselines w.r.t. the accuracy of estimating ACDE, ACME and ATE. For instance, in Table \ref{jobs1} where $\eta=10$, $n=500$, CMAVAE presents the least absolute error in all three estimations and the improvement is significant. Specifically, compared to the best baselines (i.e., LSEM and IPW), CMAVAE can reduce the error of estimating ACDE and ATE by 87.5\% and 40.0\%, respectively. The results manifest the effectiveness of the proposed framework.
    \item CMAVAE also achieves smaller standard deviations compared to other models. Its robustness, in part, benefits from the joint estimation of the unknown latent space of HC and the causal effects in CMA. The baselines, however, mostly present much larger standard deviations.
    \item For simulation parameters, strong selection of mediators ($\eta$) leads to larger bias in the estimation of ACME compared to normal selection. Larger $\alpha$ (more targets are mediated) roughly leads to larger error in ACME estimation and smaller error in ACDE estimation. Compared to baselines, CMAVAE shows more robust performance to the changes of parameters for simulation. This result further illustrates the importance as well as the challenges of decomposing the total effect in order to understand the underlying causal mechanism. 
\end{itemize}
\noindent\textbf{Varying Proxy Noise.} To further illustrate the effectiveness and robustness of CMAVAE, we introduce a new variable $p_c$, denoting the noise level of the proxies. Specifically, a larger $p_c$ indicates that we have less direct access to information of HC. Following a similar data generating procedure in \cite{louizos2017causal}, we identify HC in JOBS II as a single variable that is highly correlated with the outcome (depressive symptoms) -- PRE-DEPRESS, the pre-treatment level of depression. We then simulate proxies of PRE-DEPRESS by manually injecting noise into PRE-DEPRESS. In particular, we first simulate treatment $t_i$ using $\mat{x}_i$ and the confounder PRE-DEPRESS $\mat{z}_i$:
\begin{equation}
    t_i|\mat{x}_i,\mat{z}_i \sim \text{Bern}\big(\sigma(w_x^T\mat{x}+w_z(\frac{\mat{z}}{3}-0.3))\big),
\end{equation}
where $w_x \sim \mathcal{N}(0,0.1)$, $w_z\sim \mathcal{N}(5,0.1)$. To create the proxies for PRE-DEPRESS, we binned the samples into 3 groups based on their PRE-DPRESS values and applied one-hot encoding, which is replicated 3 times. We use three replications as previous studies have suggested that three independent views of a latent feature are what is needed to ensure its recover \cite{kruskal1976more,allman2009identifiability,anandkumar2014tensor}. We then randomly and independently flip each of these 9 binary features with probability $p_c$. We let $p_c$ vary from 0.1 to 0.5 with an increment of 0.1. $p_c=0.5$ indicates the proxies have no direct information of the confounder. 
In this experiment, we set $n=500$, $\eta=1$, and examine both $\alpha=0.5$ (50\% samples are mediated) and $\alpha=0.1$. Results averaged over 10 replications are shown in Table \ref{Confounding_Noise1}-\ref{Confounding_Noise2}.

We see from the results that CMEVAE mostly achieves the best performance when varying $p_c$. The improvement over baseline models w.r.t. ACDE and ATE is significant, see e.g., $p_c=0.5$ in Table \ref{Confounding_Noise1}(b)-(c). In addition, CMAVAE presents relatively more robust results with different proxy noise compared to other models, see, e.g., Table \ref{Confounding_Noise2}(a)-(c). This is because CMAVAE can infer a cleaner latent representation from the noisy proxies \cite{louizos2017causal}. When there are 50\% samples are mediated, i.e., $\alpha=0.5$, performance of all models roughly degrades with an increasing $p_c$. Models are more robust to increasing proxy noise when fewer samples are mediated, i.e., $\alpha=0.1$, in part because i) models can focus on the task of estimating the direct effect; and ii) the pre-treatment depression level has a stronger impact on the mediator (job search self-efficacy) than the treatment (job training or booklets) as we can see the performance degradation w.r.t. ACME when $p_c$ increases in both experiments. Overall, CMAVAE is more robust to the proxy noise than conventional methods for CMA, achieving high precision even the proxy cannot provide any useful information of HC at noise level 0.5. 
\begin{figure*}[ht!]
\centering
\begin{subfigure}{.33\textwidth}
\centering
\captionsetup{justification=centering}
  \includegraphics[width=.9\linewidth]{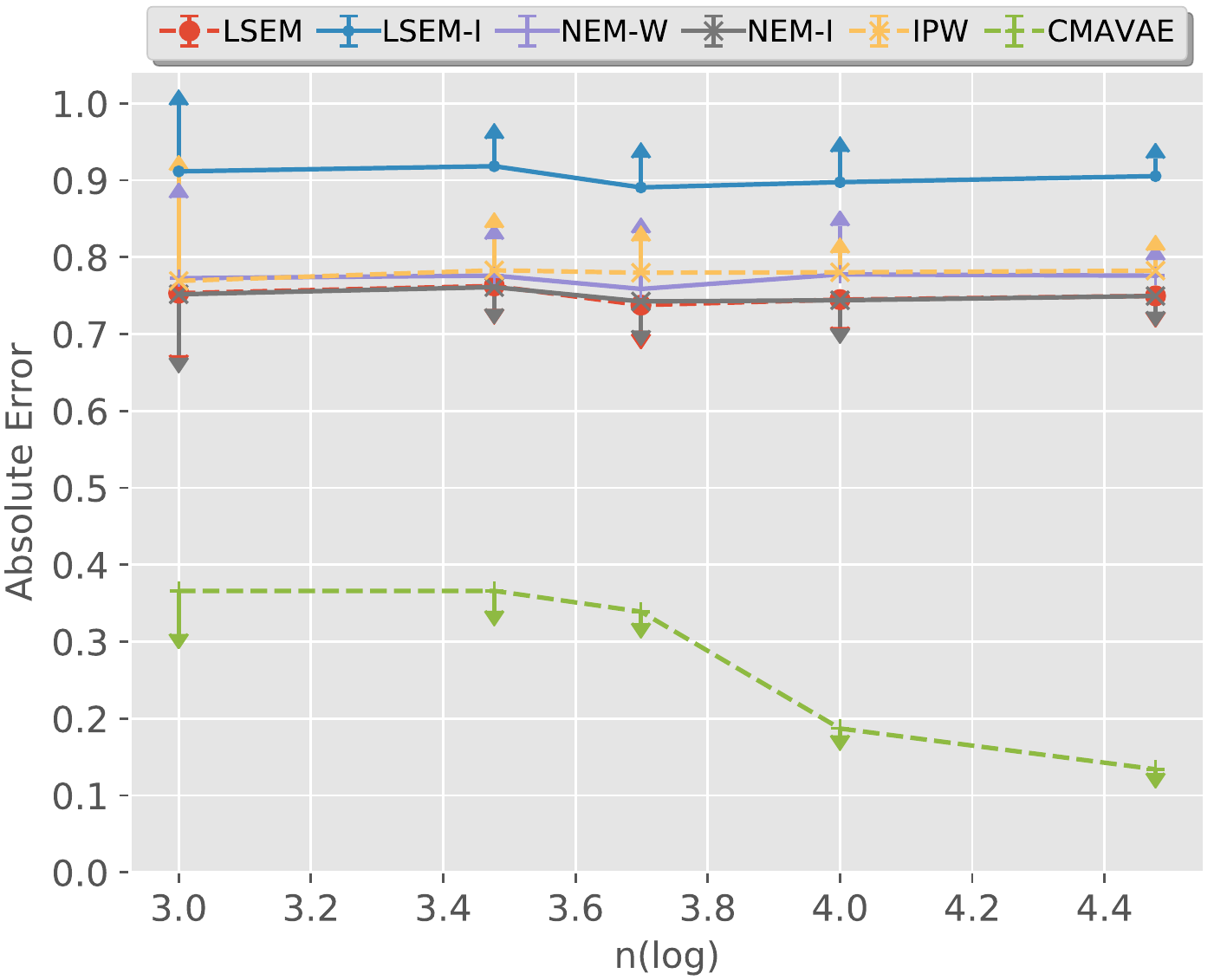}
  \caption{Estimation of ACME under treated.}
\end{subfigure}%
\begin{subfigure}{.33\textwidth}
\centering
\captionsetup{justification=centering}
  \includegraphics[width=.9\linewidth]{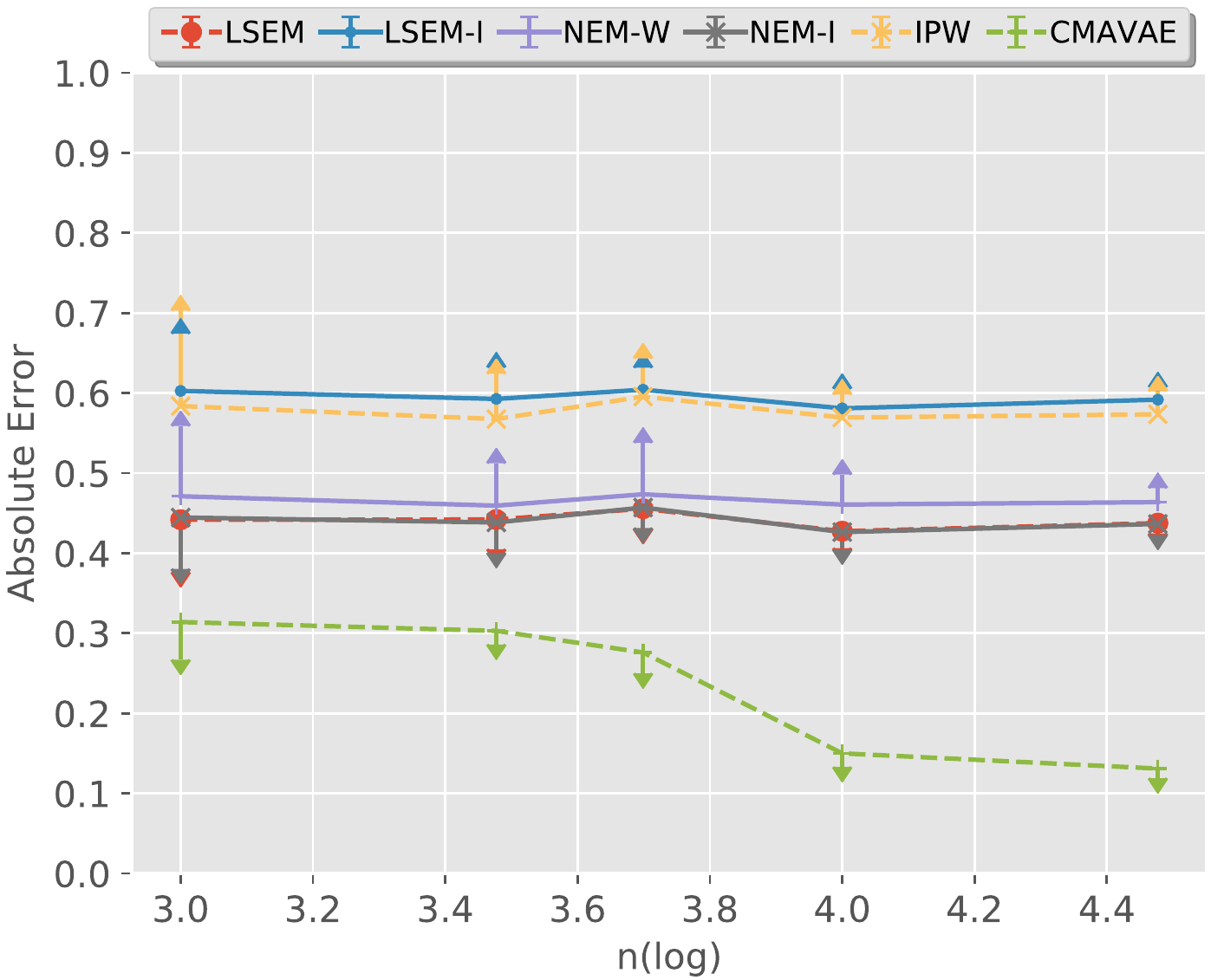}
    \caption{Estimation of ACDE under control.}
\end{subfigure}%
\begin{subfigure}{.33\textwidth}
\centering
\captionsetup{justification=centering}
  \includegraphics[width=.9\linewidth]{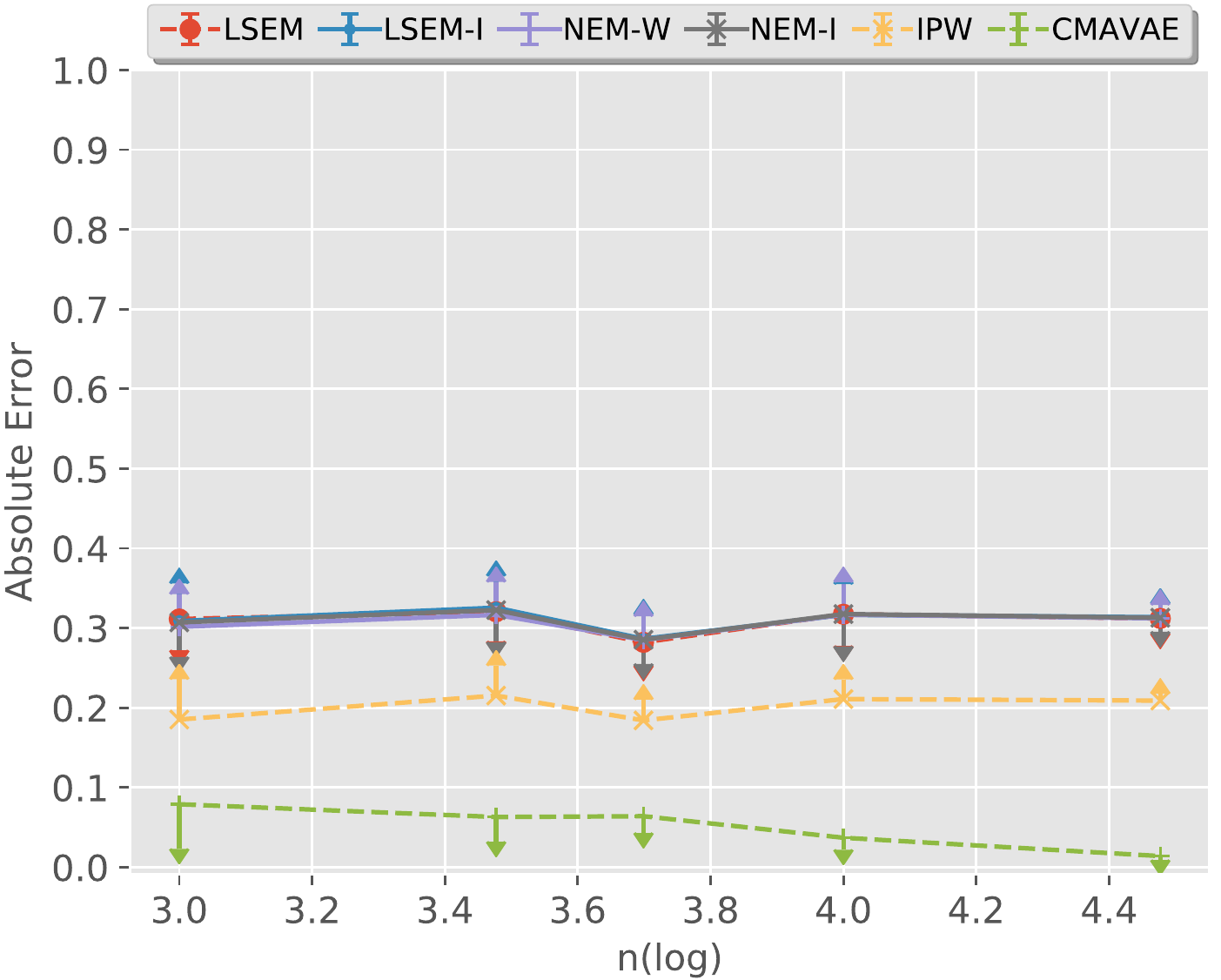}
    \caption{Estimation of ATE.}
\end{subfigure}
\caption{Absolute errors of the estimated ACME, ACDE, ATE on synthetic samples simulated from the data generative process (Eq.~\eqref{DGP}).}
\label{toy}
\end{figure*}
\subsection{Synthetic Experiments}
Suppose that $\mat{X}$ are the attributes of customers, $\mat{M}$ is the engagement of users in organic search, $T$ denotes whether a customer is using the new recommendation module $(T=1)$ or not $(T=0)$, $Y$ is the conversion rate, and $\mat{Z}$ is the HC. We define $\mat{Z}$, $\mat{M}$, $\mat{X}$, $T$ and $Y$ as
\begin{equation}
\begin{split}
    \mat{z}_i\sim \text{Bern}(0.5); \\
    \mat{x}_i|\mat{z}_i \sim \mathcal{N}(\mat{z}_i,25\mat{z}_i+9(1-\mat{z}_i));\\
    t_i|\mat{z}_i \sim \text{Bern}(0.75\mat{z}_i+0.25(1-\mat{z}_i));\\
    \mat{m}_i|\mat{z}_i,t_i \sim 0.5\mat{z}_i+0.5t_i\bm{\kappa}_{i}+\mat{e}_1;\\
    y_i|\mat{z}_i,t_i,\mat{m}_i \sim \mat{c}\mat{z}_i+t_i+\mat{m}_i+0.5t_i\mat{m}_i+\mat{e}_2,
\end{split}
\label{DGP}
\end{equation}
where $\bm{\kappa}_i=\frac{1}{1+\exp(-(1+0.2\mat{z}_i))}$ models the influence of using new recommendation module on user engagement through the HC $\mat{z}_i$. We let $\mat{c}\sim \mathcal{N}(0,1)$ be the scaling factor and $\mat{e}_1,\mat{e}_2\sim \mathcal{N}(0,1)$ represent the Gaussian noise. This data generating process explicitly introduces HC between $t$ and $\mat{m}$, $t$ and $y$, and $\mat{m}$ and $y$ as they all hinge on the mixture assignment $\mat{z}$ for $\mat{x}$. As the sequential ignorability assumption has been violated, we expect that baselines may not accurately estimate ACME, ACDE and ATE. We model $\mat{z}$ as a 5-dimensional continuous variable following the Gaussian distribution to investigate the robustness of CMAVAE w.r.t. model misspecification. We evaluate across sample sizes $n \in \{1000, 3000,5000, 10000, 30000\}$ and present the mean and standard deviation of each approach in Fig. \ref{toy}. 

\noindent\textbf{Results.} CMAVAE achieves significantly less error and variance than the baseline models across various settings although we purposely misspecified the latent model. 
% {\color{blue} reviewers may wonder what if the prior of $z$ is correct.} 
Specifically, the relative improvement over baselines is the largest for estimating ACME, smaller for ACDE, and the least for ATE. This implies that estimated ACDE and ACME deviate from the ground truth in opposite directions, yielding a total effect that is closer to the ground truth.
Additionally, the improvement w.r.t. the averages of estimated effects and the corresponding standard deviations becomes more significant when the sample size increases, showing the promises of using big-data for causal inference in the presence of proxies for HC \cite{louizos2017causal}.

% In summary, CMAVAE outperforms the standard approaches to CMA w.r.t. both effectiveness and robustness of estimating ACDE under control, ACME under treated and ATE in the presence of HC. The conclusion applies to both synthetic and semi-synthetic datasets generated from real-world data. This suggests the benefits of using proxy and deep latent-variable models to control for HCs.
\subsection{Application in Causal Fairness Analysis}
\begin{figure}
    \centering
      \includegraphics[width=.45\linewidth]{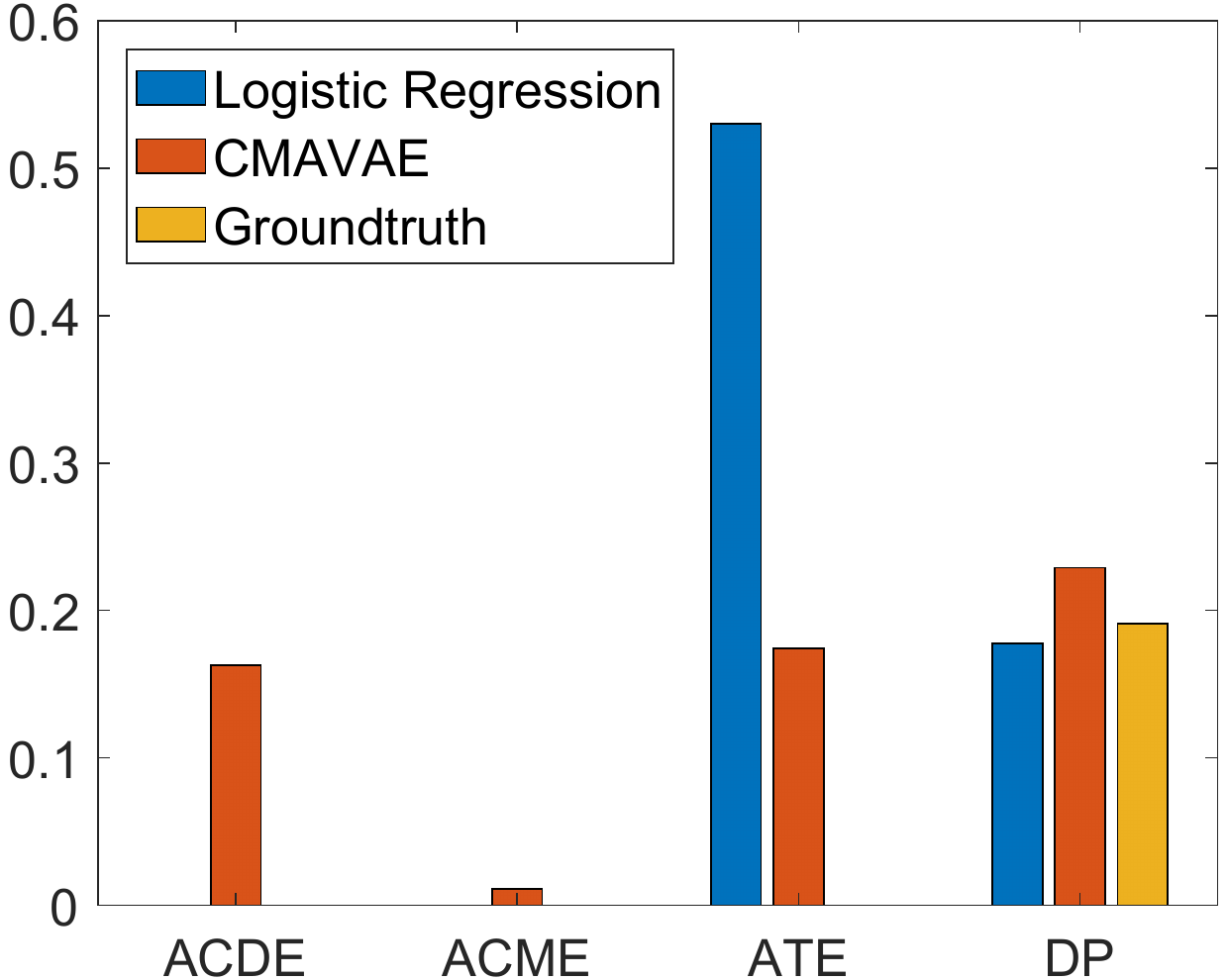}
    \caption{Comparisons of statistical and causal fairness analyses.}
\label{fairness}
\end{figure}

In addition to the simulation-based evaluation, we apply CMAVAE to the real-world example described in Example 2. This experiment illustrates how CMAVAE can help detect and interpret discrimination in machine learning from a causal perspective. In particular, we partition discrimination into two components -- \textit{direct} and \textit{indirect} discrimination \cite{national2004measuring}. We use the benchmark UCI Adult dataset \cite{Dua:2019} where the sensitive attribute \textit{Gender} is the treatment, \textit{Occupation} the mediator, and \textit{Income} ($>50K, \leq 50K$) the outcome \cite{zhang2018fairness}. All other demographic covariates (e.g., marital status) are used as the proxies of HC. In comparison, we employ the statistical definition of fairness -- demographic disparity (DP) \cite{dwork2012fairness} -- and Logistic Regression as the classifier for its interpretability and simplicity. Dealing in correlations, DP requires the outcome to be independent of the sensitive attributes, i.e., $P(\hat{Y}|T=0)=P(\hat{Y}|T=1)$. We report results for ACME, ACDE, ATE, and DP in Fig. \ref{fairness} to show a bird's-eye view of (1) how gender might affect the annual income directly (ACDE) and indirectly (ACME); (2) the differences of the discrimination measure between statistical and causal fairness (ATE and DP). Note that ``ATE'' under statistical fairness is the estimated coefficient of gender in the classifier and it cannot be split into direct and indirect effects. For CMAVAE, ATE=0 denotes non-discrimination. As we cannot know the ground-truth causal effects of gender on income, we only show the ground truth of DP calculated with true $Y$. 

We observe that (1) both the ``ATE'' of gender for logistic regression and the ATE for CMAVAE are positive, indicating that there is potential discrimination against female employees. We can reach the similar conclusion from the results of DP; (2) based on CMAVAE results, a small part of the discrimination can be attributed to the indirect influence from occupation which is also affected by gender. We cannot get such information from statistical fairness; and (3) under statistical fairness notion, logistic regression presents closer DP to the ground truth. As DP relies on the accuracy of inferred $Y$, a better results from logistic regression is expected because highly predictive yet correlated information were removed from CMAVAE. Of particular interest is the contradictory results for ATE and DP: under causal fairness measure, ``ATE'' of gender for logistic regression is significantly larger than ATE for CMAVAE whereas the results are opposite under statistical fairness measure DP. This signals the inherent differences between causal and statistical fairness notions. Future research on the comparisons is warranted. 
\section{Conclusions \& Future Work}
This work studies an important causal inference problem that seeks to break down the total effect of a treatment into direct and indirect causal effects. When direct intervention on the mediator is not possible, CMA shows its potentials to reveal the underlying causal mechanism involved with treatment, mediator, and outcome. When applied to observational data, CMA relies on the sequential ignorability assumption -- confounder bias can be controlled for by the observable covariates -- that is often unverifiable in practice. This work circumvents the strong assumption and studies CMA in the presence of HC. To achieve the goal, we approximate HC by incorporating proxy variables that are easier to measure. The proposed approach CMAVAE draws connections between conventional CMA and recent advances in deep latent-variable models to simultaneously estimate HC and causal effects. Experimental results show the efficacy and potential applications of our approach.

This study opens promising future directions. First, we plan to investigate the influence of proxy variables on estimating ACME and ACDE separately. The empirical evaluations in Sec. 5 suggests that there might be a trade-off between these two tasks. We may further design a weighting strategy to balance between ACME and ACDE, increasing model's flexibility in solving different tasks. In addition, we would like to extend this work to settings where multiple mediators are of particular interest. To apply CMAVAE into real-life applications such as clinical trials and recommendation systems, one needs to be aware of the scientific context of that specific scenario in order to maximize the performance of CMAVAE. 
\section*{Acknowledgements} This work is supported by ONR N00014-21-1-4002 and ARO W911NF\\2110030. The views, opinions and/or findings expressed are the authors' and should not be interpreted as representing the official views or policies of the ARO or ONR.
\bibliographystyle{ACM-Reference-Format}
\bibliography{sample-base}

\appendix
\section{Proof of Theorem 1}
\noindent\textit{Proof.} The key is to show that $p(y|\mat{X},\mat{M}(do(T=t')),t)$ is identifiable. We have 
\begin{gather}
\begin{split}
    p(y|\mat{X}, \mat{M}(do(T=t')),t)=\\
    \int_{\mat{Z}}p(y|\mat{X},\mat{M}(do(T=t')),t,\mat{Z})p(\mat{Z}|\mat{X},\mat{M}(do(T=t')),t)d\mat{Z}\\
    % \stackrel{do}
    {=}\int_{\mat{Z}}p(y|\mat{X},\mat{M}(t'),t,\mat{Z})p(\mat{Z}|\mat{X},\mat{M}(t'),t)d\mat{Z},
\end{split}
\label{theoremproof}
\end{gather}
where the second equality holds by applying the \textit{do}-calculus rule to the causal graph in Fig.1. The final expression in Eq. (\ref{theoremproof}) can be identified from the distribution $p(\mat{Z},\mat{X},\mat{M},t,y)$. Similarly, we can prove that $p(y|\mat{X},\mat{M}(t),do(T=t'))$ for ACDE is also identifiable given the joint probability distribution $p(\mat{Z},\mat{X},\mat{M},t,y)$:
\begin{equation}
\begin{split}
    \overline{\zeta}(t)&:=\mathbb{E} [CDE(\mat{x},t)], \text{with}\\
     CDE(\mat{x},t)&:=\mathbb{E}[y|\mat{X}=\mat{x}, do(t'=1), \mat{M}(T=t))]\\
     &-\mathbb{E}[y|\mat{X}=\mat{x}, do(t'=0),  \mat{M}(T=t)], \quad t=0,1.
\end{split}
\end{equation}
\begin{table}
\caption{Details of the parameter settings in proposed models for both simulation data and real-world datasets.}
\begin{tabular}{|l|l|l|l|}
\hline
Parameter              & Simulation & JOBS II & Adult \\ \hline\hline
Reps                 & 10            &10          & 1   \\ \hline
Epoch                  & 100          & 100      & 150        \\ \hline
Embed\_Size             & 5           & 10    & 10      \\ \hline
Layer\_Size            & 100           & 100     & 100    \\ \hline
Batch\_Size            & 100         & 32   & 1024     \\ \hline
$lr$       & 1e-4        & 1e-6    & 1e-5       \\ \hline
n\_layers              & 3            & 5   & 2       \\ \hline
$\lambda$ & 1e-4         & 1e-3   & 1e-3      \\ \hline
$L$        & 100          &  100  &  1000         \\ \hline
\end{tabular}
\label{parameters}
\end{table}
\section{CMAVAE in Binary Cases}
With binary mediator and outcome, we first define
\begin{gather}
    p(\mat{z}_i)=\prod_{j=1}^{D_z}\mathcal{N}(\mat{z}_{ij}|0,1); \quad p(\mat{x}_i|\mat{z}_i)=\prod_{j=1}^{D_x}p(\mat{x}_{ij}|\mat{z}_i); \\ p(t_i|\mat{z}_i)=\text{Bern}(\sigma(f_1(\mat{z}_i))),
\end{gather}
CMAVAE can then be formulated as follows:
\begin{gather}
    p(\mat{m}_i|\mat{z}_i,t_i)=\text{Bern}(\pi=\hat{\pi}_{1i});\\
     p(y_i|\mat{m}_i,\mat{z}_i,t_i)= \text{Bern}(\pi=\hat{\pi}_{2i}),
\end{gather}
where 
\begin{gather}
    \hat{\pi}_{1i}=\sigma(t_if_2(\mat{z}_i)+(1-t_i)f_3(\mat{z}_i));\\
    \hat{\pi}_{2i}=\sigma(t_if_4(\mat{z}_i\circ \mat{m}_i)+(1-t_i)f_5(\mat{z}_i\circ \mat{m}_i)).
\end{gather}
Then the posterior over $\mat{Z}$ can be approximated by 
\begin{gather}
    q(\mat{z}_i|\mat{x}_i,\mat{m}_i,y_i,t_i)=\prod_{j=1}^{D_z}\mathcal{N}(\mu_j=\overline{\mu}_{ij},\sigma^2_j=\overline{\sigma}^2_{ij});\\
    \overline{\bm{\mu}}_i=t_i\bm{\mu}_{t=0,i}+(1-t_i)\bm{\mu}_{t=1,i};\\ \overline{\bm{\sigma}}_i^2=t_i\bm{\sigma}^2_{t=0,i}+(1-t_i)\bm{\sigma}^2_{t=1,i};\\
    \bm{\mu}_{t=0,i},\bm{\sigma}^2_{t=0,i}=g_1(\mat{x}_i\circ y_i \circ \mat{m}_i);\\
     \bm{\mu}_{t=1,i},\bm{\sigma}^2_{t=1,i}=g_2(\mat{x}_i\circ y_i \circ \mat{m}_i), 
\end{gather}
The objective function of VAEs is then:
\begin{gather}
\begin{split}
    \mathcal{L}=\sum_{i=1}^n\mathbb{E}_{q(\mat{z}_i|\mat{x}_i,t_i,\mat{m}_i,y_i)}[\log p(\mat{z}_i)+\log p(\mat{x}_i|\mat{z}_i)+\log p(t_i|\mat{z}_i)\\
    +\log p(\mat{m}_i|\mat{z}_i,t_i)
    +\log p(y_i|\mat{m}_i,\mat{z}_i,t_i)-\log q(\mat{z}_i|\mat{x}_i,\mat{m}_i,y_i,t_i)].
\end{split}
\end{gather}
However, in the Encoder, we still need to infer the treatment assignment $t$, mediator $\mat{m}$, and outcome $y$ for new subjects before inferring the posterior distribution over $\mat{z}$. As a result, we introduce three auxiliary distributions that seek to predict $t_i, \mat{m}_i, y_i$ for new samples with covariates $\mat{x}_i$. They are
\begin{equation}
    q(t_i|\mat{x}_i)=\text{Bern}(\pi=\sigma(g_3(\mat{x}_i)));
\end{equation}
\begin{equation}
\begin{aligned}
q(\mat{m}_i|\mat{x}_i,t_i)&=\text{Bern}(\pi=\overline{\pi}_{1i}),
\end{aligned}
\end{equation}
where 
\begin{gather}
    \overline{\pi}_{1i}=\sigma(t_ig_4(\mat{x}_i)+(1-t_i)g_5(\mat{x}_i)).
\end{gather}
Similarly, we have 
\begin{equation}
\begin{aligned}
 q(y_i|\mat{x}_i,\mat{m}_i,t_i)&=\text{Bern}(\pi=\overline{\pi}_{2i}),
\end{aligned}   
\end{equation}
where 
\begin{gather}
    \overline{\pi}_{2i}=\sigma(t_ig_{6}(\mat{x}_i\circ \mat{m}_i)+(1-t_i)g_7(\mat{x}_i\circ \mat{m}_i)).
\end{gather}
% We add these auxiliary terms to the objective function in order to ``learn'' the parameters of $q(t|\mat{x})$, $q(\mat{m}|\mat{x},t)$ and $q(y|\mat{x},\mat{m},t)$.
The final loss is defined as 
\begin{equation}
    \begin{split}
        \mathcal{F}&=\mathcal{L}+\sum_{i=1}^n (\log q(\mat{m}_i=\mat{m}_i^*|\mat{x}_i^*,t_i^*)+\\
        \log q(t_i&=t_i^*|\mat{x}_i^*)+\log q(y_i=y_i^*|\mat{x}_i^*,\mat{m}_i^*,t_i^*)),
    \end{split}
\end{equation}
where $y_i^*,\mat{x}_i^*,\mat{m}_i^*,t_i^*$ are the observed values for the outcome, input, mediator and treatment random variables in the training set. 
\section{Reproducibility}
In this section, we provide more details of the experimental setting and configuration for reproducibility purpose.

Our proposed models were implemented in Python library Tensorflow \cite{abadi2016tensorflow} and Edward \cite{tran2016edward}. Code for data simulation and all baselines is written in R. We detail the parameter settings of the proposed models for simulation data, JOBS II data, and the Adult data for causal fair analysis in Table \ref{parameters}. The descriptions of the major parameters are introduced below:
\begin{itemize}
    \item Reps: the number of replications each set of experiments runs. The parameters used are fixed, but each replication of the generated data can be different.
    \item Epoch: one Epoch is when an entire dataset is passed forward and backward through the neural network only once.
    \item Embed\_Size: the dimensions of the latent variable $Z$. 
    \item Layer\_Size: the output size of every layer.
    \item Batch\_Size: total number of training examples present in a single batch.
    \item $lr$: the learning rate.
    \item n\_layers: the number of hidden layers. 
    \item $\lambda$: the hyperparameter for weight decay.
    \item $L$: the number of samples drawn from the posterior distribution to estimate mediator and outcome.
\end{itemize}
\end{document}